\documentclass[11pt]{article}
\usepackage{coling2016}
\usepackage{times}
\usepackage{url}
\usepackage{latexsym}

\usepackage[utf8x]{inputenc} 
\usepackage[english]{babel}
\usepackage[disable]{todonotes} 
\usepackage{amsmath}
\usepackage{amsfonts}
\usepackage{amssymb}
\usepackage{booktabs}

\newcommand{\Sec}[1]{Section~\ref{#1}}

\newcommand{\Fig}[1]{Figure~\ref{#1}}
\newcommand{\Tab}[1]{Table~\ref{#1}}

\renewcommand{\vec}[1]{\mathbf{#1}}
\newcommand{\mat}[1]{{#1}} 

\def\weights{\mat{W}}
\def\bias{\vec{b}}
\def\allparams{\mat{\Omega}}
\def\vocab{V}
\def\doc{\mathcal{D}}
\def\corpus{\mathcal{C}}
\def\trainingset{\mathcal{I}}
\def\word{w}
\def\pos{n}
\def\onehot{\vec{v}}
\def\lembedding{{x}}
\def\lstmparams{\mat{\Theta}}

\def\lstmsinput{\vec{x}}

\def\leftlstmsoutput{\vec{h}^L}
\def\rightlstmsoutput{\vec{h}^R}

\def\hiddenlayer{\vec{a}}
\def\sensepredict{\vec{y}}

\def\gaussnoise{\sigma_i}

\def\SEtwo{SE2}
\def\SEthree{SE3}
\def\dropword{dropword}
\def\dropout{dropout}
\def\worddropout{word dropout}
\def\glove{GloVe}
\def\SEtwoBest{100JHU(R)}
\def\SEthreeBest{htsa3}
\def\IMS{IMS+adapted CW}
\def\oursystem{BLSTM}

\def\dropword{dropword}
\def\Dropword{Dropword}

\def\randomwordorder{randomized word order}

\title{Word Sense Disambiguation using a Bidirectional LSTM}

\newcommand*\samethanks[1][\value{footnote}]{\footnotemark[#1]}
\author{Mikael K{\aa}geb{\"a}ck\thanks{~Authors contributed equally.}, Hans Salomonsson\samethanks[1]\\
Computer Science \& Engineering, Chalmers University of Technology \\
SE-412 96, G\"{o}teborg, Sweden\\
  {\tt \{kageback,hans.salomonsson\}@chalmers.se} }
  
\date{}

\begin{document}
\maketitle
\begin{abstract}
In this paper we present a clean, yet effective, model for \emph{word sense disambiguation}. 
Our approach leverage a \emph{bidirectional long short-term memory} network which is shared between all words. This enables the model to share statistical strength and to scale well with vocabulary size.
The model is trained \emph{end-to-end}, directly from the raw text to sense labels, and makes effective use of word order. 
We evaluate our approach on two standard datasets, using identical hyperparameter settings, which are in turn tuned on a third set of held out data. 
We employ no external resources (e.g. knowledge graphs, part-of-speech tagging, etc), language specific features, or hand crafted rules, but still achieve statistically equivalent results to the best \emph{state-of-the-art} systems, that employ no such limitations.  

\end{abstract}
\section{Introduction}
\blfootnote{This work is licenced under a Creative Commons 
           Attribution 4.0 International License.
           License details:
           \url{http://creativecommons.org/licenses/by/4.0/}}
Words are in general ambiguous and can have several related or unrelated meanings depending on context. For instance, the word \emph{rock} can refer to both a stone and a music genre, but in the sentence "Without the guitar, there would be no \emph{rock} music" the sense of \emph{rock} is no longer ambiguous. The task of assigning a word token in a text, e.g. \emph{rock}, to a well defined word sense in a lexicon is called \emph{word sense disambiguation} (WSD). 
From the \emph{rock} example above it is easy to see that the context surrounding the word is what disambiguates the sense. However, it may not be so obvious that this is a difficult task. To see this, consider instead the phrase "Solid rock" where changing the order of words completely changes the meaning, or "Hard rock crushes heavy metal" where individual words seem to indicate stone but together they actually define the word token as music. With this in mind, our thesis is that to do WSD well we need to go beyond \emph{bag of words} and into the territory of sequence modeling. 

Improved WSD would be beneficial to many natural language processing (NLP) problems, e.g. machine translation \cite{vickrey2005word}, information Retrieval, information Extraction \cite{Navigli:09}, and sense aware word representations \cite{neelakantan2015efficient,kaageback2015neural,nietopina2015,bovi2015knowledge}. However, though much progress has been made in the area, many current WSD systems suffer from one or two of the following deficits. (1) Disregarding the order of words in the context which can lead to problems as described above. (2) Relying on complicated and potentially language specific hand crafted features and resources, which is a big problem particularly for resource poor languages.
We aim to mitigate these problems by (1) modeling the sequence of words surrounding the target word, and (2) refrain from using any hand crafted features or external resources and instead represent the words using real valued vector representation, i.e. word embeddings. Using word embeddings has previously been shown to improve WSD \cite{taghipour2015semi,johansson2015combining}. However, these works did not consider the order of words or their operational effect on each other.

\subsection{The main contributions of this work include:}
\begin{itemize}
\item A purely learned approach to WSD that achieves results on par with state-of-the-art resource heavy systems, employing e.g. knowledge graphs, parsers, part-of-speech tagging, etc. 
\item Parameter sharing between different word types to make more efficient use of labeled data and make full vocabulary scaling plausible without the number of parameters exploding.
\item Empirical evidence that highlights the importance of word order for WSD.
\item A WSD system that, by using no explicit window, is allowed to combine local and global information when deducing the sense.
\end{itemize}
\section{Background}
In this section we introduce the most important underlying techniques for our proposed model. 
\subsection{Bidirectional LSTM}
\label{sec:BLSTM}
\emph{Long short-term memory} (LSTM) is a gated type of  \emph{recurrent neural network} (RNN). LSTMs were introduced by \newcite{hochreiter1997long} to enable RNNs to better capture long term dependencies when used to model sequences. This is achieved by letting the model copy the state between timesteps without forcing the state through a non-linearity. The flow of information is instead regulated using multiplicative gates which preserves the gradient better than e.g. the logistic function.
%
The bidirectional variant of LSTM, (BLSTM) \cite{graves2005framewise} is an adaptation of the LSTM where the state at each time step consist of the state of two LSTMs, one going left and one going right. For WSD this means that the state has information about both preceding words and succeeding words, which in many cases are absolutely necessary to correctly classify the sense. 
\subsection{Word embeddings by \glove}
Word embeddings is a way to represent words as real valued vectors in a semantically meaningful space. \emph{Global Vectors for Word Representation} (\glove), introduced by \newcite{pennington2014glove} is a hybrid approach to embedding words that combine a log-linear model, made popular by \newcite{mikolov2013linguistic}, with counting based co-occurrence statistics to more efficiently capture global statistics. Word embeddings are trained in an unsupervised fashion, typically on large amounts of data, and is able to capture fine grained semantic and syntactic information about words. These vectors can subsequently be used to initialize the input layer of a neural network or some other NLP model.
\section{The Model}
\begin{figure}
 \centering
  \def\layersep{1.2cm} 
\def\timestepsep{2.1cm} 
\def\layerwidth{1.0cm}
\def\layerheight{0.5cm}
\def\outlayer{4.0cm}
\def\one{1.0}

\begin{tikzpicture}[shorten >=1pt,->,draw=black!90, node distance=\layersep]

\tikzstyle{every pin edge}=[<-,shorten <=1pt]	
\tikzstyle{layer}=[	rectangle,
					fill=black!25,
                    minimum width=\layerwidth, 
                    minimum height=\layerheight]
\tikzstyle{LRNN}=[layer, fill=red!50];
\tikzstyle{RRNN}=[layer, fill=blue!50];

\tikzstyle{annot} = [text width=1em, text centered]

\foreach \name / \i in {1,...,3}
	\path[xshift=\timestepsep-0.5*\layerwidth]
        node[LRNN, pin=below:$ \lstmsinput_{\pos-\i} $] 
        	(L-\name) at (-\i*\timestepsep+\one ,0) {};

\path[xshift=\timestepsep-0.5*\layerwidth]
	node[LRNN, pin=below:$ \lstmsinput_{0} $] 
		(L-0) at (-4*\timestepsep+\one ,0) {};

\foreach \dest / \source in {1/2,2/3}
    	\path (L-\source) edge (L-\dest);
 \node[annot,left of=L-3, node distance=1.0*\layerwidth] {...};

\foreach \name / \i in {1,...,3}
	\path[xshift=-\timestepsep+0.5*\layerwidth]
        node[RRNN, pin=below:$ \lstmsinput_{\pos+\i} $] 
        	(R-\name) at (\i*\timestepsep,0) {};

\path[xshift=-\timestepsep+0.5*\layerwidth]
 	node[RRNN, pin=below:$ \lstmsinput_{|\doc|} $] 
        (R-D) at (4*\timestepsep,0) {};

\foreach \dest / \source in {1/2,2/3}
    	\path (R-\source) edge (R-\dest);
 \node[annot,right of=R-3, node distance=1.0*\layerwidth] {...};
 
 
\node[layer,pin=below:] (H) at (0,\layersep) {};
\node[annot,yshift=0.0cm,right of=H, node distance=0.8cm] {\(\hiddenlayer\)};

\foreach \x / \y in {0.1/2.1,0.2/2.2}
	\node[layer,fill=black!15] at (\x,\y*\layersep){};

\node[layer,pin={[pin edge={->}]above:\(\sensepredict(\pos)\)},pin={[shorten >= 5pt]below:}] (O) at (0,2*\layersep) {};

\node[annot,yshift=0.3cm,right of=O, node distance=1.2cm] {};

    
\end{tikzpicture}
  \caption{A BLSTM centered around a word at position $ \pos $. Its output is fed to a neural network sense classifier consisting of one hidden layer with linear units and a softmax. The softmax selects the corresponding weight matrix and bias vector for the word at position $n$. }.
  \label{fig:model}
\end{figure}
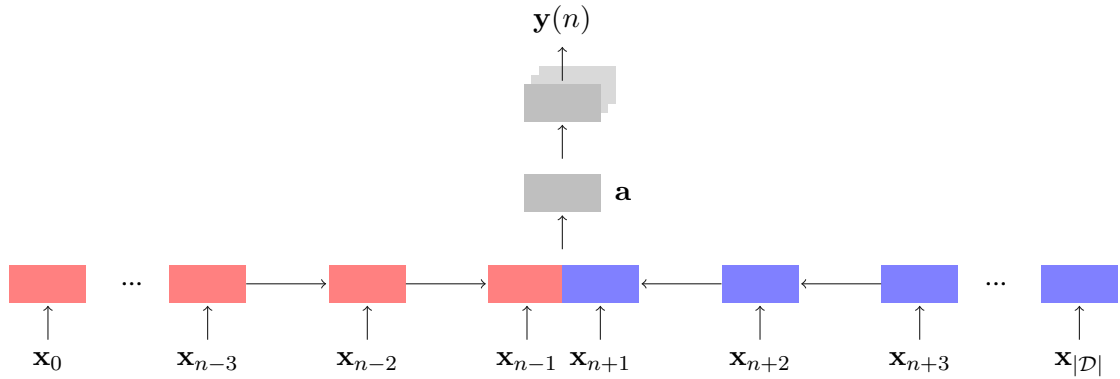

Given a document and the position of the target word, i.e. the word to disambiguate, the model computes a probability distribution over the possible senses corresponding to that word. The architecture of the model, depicted in \Fig{fig:model}, consist of a softmax layer, a hidden layer, and a BLSTM. See \Sec{sec:BLSTM} for more details regarding the BLSTM. The BLSTM and the hidden layer share parameters over all word types and senses, while the softmax is parameterized by word type and selects the corresponding weight matrix and bias vector for each word type respectively. This structure enables the model to share statistical strength across different word types while remaining computationally efficient even for a large total number of senses and realistic vocabulary sizes.

\subsection{Model definition}
The input to the BLSTM at position \(\pos\) in document \(\doc\) is computed as
\[
\lstmsinput_\pos=\weights^{\lembedding} \onehot(\word_\pos),\pos \in \{1,\dots,|\doc|\}.
\]
Here, \(\onehot(\word_\pos)\) is the \emph{one-hot} representation of the word type corresponding to \(\word_\pos\in\doc\). A one-hot representation is a vector with dimension $\vocab$ consisting of $|\vocab|-1$ zeros and a single one which index indicate the word type. This will have the effect of picking the column from \(\weights^{\lembedding}\) corresponding to that word type. The resulting vector is referred to as a word embedding. Further, \(\weights^{\lembedding}\) can be initialized using pre-trained word embeddings, to leverage large unannotated datasets. In this work GloVe vectors are used for this purpose, see \Sec{sec:experimentalsettings} for details.

The model output, 
\[
\sensepredict(\pos) = \mathrm{softmax}(\weights^{ay}_{\word_\pos} \hiddenlayer + \bias^{ay}_{\word_\pos}),
\]
is the predicted distribution over senses for the word at position \(\pos\), where $\weights^{ay}_{\word_\pos}$ and $\bias^{ay}_{\word_\pos}$ are the weights and biases for the softmax layer corresponding to the word type at position $\pos$. Hence, each word type will have its own softmax parameters, with dimensions depending on the number of senses of that particular word. 
Further, the hidden layer \(\hiddenlayer\) is computed as 
\[
\hiddenlayer = \weights^{ha} [\leftlstmsoutput_{\pos-1};\rightlstmsoutput_{\pos+1}] + \bias^{ha}
\] 
where $[\leftlstmsoutput_{\pos-1};\rightlstmsoutput_{\pos+1}]$ is the concatenated outputs of the right and left traversing LSTMs of the BLSTM at word $n$. $\weights^{ha}$ and $\bias^{ha}$ are the weights and biases for the hidden layer.

%
%

\paragraph{Loss function}
The parameters of the model,
\(\allparams=\{ \weights^{\lembedding}, \lstmparams_{BLSTM}, \)
\(\weights^{ha}, \bias^{ha}, \{\weights^{ay}_{\word}, \bias^{ay}_{\word} \}_{\forall \word \in \vocab}, 
\},\)
are fitted by minimizing the cross entropy error 
\[
L(\allparams)=-\sum_{i\in \trainingset} \sum_{j \in S(\word_i)}t_{i,j} \log y_j(i)
\]
over a set of sense labeled tokens with indices \(\trainingset \subset \{1,\dots,|\corpus|\} \) within a training corpus \(\corpus\), each labeled with a target sense \(\vec{t}_i,\forall i \in \trainingset\).  

\subsection{\Dropword}
\emph{\Dropword{}} is a regularization technique very similar to \emph{\worddropout{}} introduced by \newcite{iyyer2015deep}. Both methods are word level generalizations of \dropout{}~\cite{srivastava2014dropout} but in \worddropout{} the word is set to zero while in \dropword{} it is replaced with a \emph{\textless dropped\textgreater} tag. The tag is subsequently treated just like any other word in the vocabulary and has a corresponding word embedding that is trained. This process is repeated over time, so that the words dropped change over time. The motivation for doing \dropword{} is to decrease the dependency on individual words in the training context. This technique can be generalized to other kinds of sequential inputs, not only words. 
\section{Experiments}
To evaluate our proposed model we perform the \emph{lexical sample task} of SensEval~2~(\SEtwo) \cite{kilgarriff2001english} and  SensEval~3~(\SEthree) \cite{mihalcea2004senseval}, part of the SensEval \cite{kilgarriff2000introduction} workshops organized by \emph{Special Interest Group on the Lexicon} at ACL. For both instances of the task training and test data are supplied, and the task consist of disambiguating one indicated word in a context. The words to disambiguate are sampled from the vocabulary to give a range of low, medium and high frequency words, and a gold standard sense label is supplied for training and evaluation.
\subsection{Experimental settings}

The hyperparameter settings used during the experiments, presented in \Tab{tbl:experimentalsettings}, were tuned on a separate validation set with data picked from the \SEtwo{} training set. The source code, implemented using \emph{TensorFlow}~\cite{tensorflow2015-whitepaper}, has been released as open source\footnote{Source for all experiments is available at: \url{https://bitbucket.org/salomons/wsd}}.

\label{sec:experimentalsettings}
\begin{table}[tbh]
\centering
\begin{tabular*}{0.9\textwidth}{@{\extracolsep{\fill} } lcr }
\toprule 
\bf Hyperparameter & \bf Range searched & \bf Value used \\
\midrule
Embedding size & \{100, 200\} & $100$ \\
BLSTM hidden layer size & [50, 100] & $2*74$ \\ 
Dropout on word embeddings $\lstmsinput_\pos$ & [0, 50\%] & $50$\% \\
Dropout on the LSTM output $[\leftlstmsoutput_{\pos-1};\rightlstmsoutput_{\pos+1}]$ & [0, 70\%] & $50$\% \\
Dropout on the hidden layer \(\hiddenlayer\) & [0, 70\%] & $50$\% \\
Dropword & [0, 20\%] & $10$\% \\
Gaussian noise added to input & [0, 0.4] & \(\sim\mathcal{N}(0,0.2\gaussnoise)\) \\ 
\midrule
Optimization algorithm & - & Stochastic gradient descent\\
Momentum & - & 0.1 \\
Initial learning rate & - & $2.0$ \\ 
Learning rate decay & - & $0.96$ \\
Embedding initialization & - & \glove{} \\
Remaining parameters initialized & - & $\in \mathcal{U}(-0.1, 0.1)$ \\
\bottomrule
\end{tabular*}
\caption{Hyperparameter settings used for both experiments and the ranges that were searched during tuning. "-" indicates that no tuning were performed on that parameter.}
\label{tbl:experimentalsettings}
\end{table}

\paragraph{Embeddings} The embeddings are initialized using a set of freely available\footnote{The employed \glove{} vectors are available for download at: \url{http://nlp.stanford.edu/projects/glove/}} \glove{} vectors trained on Wikipedia and Gigaword. Words not included in this set are initialized from $\mathcal{N}(0, 0.1)$.
To keep the input noise proportional to the embeddings it is scaled by $\gaussnoise$ which is the standard deviation in embedding dimension $i$ for all words in the embeddings matrix, $\weights^{\lembedding}$. $\gaussnoise$ is updated after each weight update.

\paragraph{Data preprocessing} 
The only preprocessing of the data that is conducted is replacing numbers with a $<number>$ tag. This result in a vocabulary size of $|\vocab| = 50817$ for \SEtwo{} and $|\vocab| = 37998$ for \SEthree. Words not present in the training set are considered unknown during test. 
Further, we limit the size of the context to max 140 words centered around the target word to facilitate faster training.
\subsection{Results}
The results of our experiments and the state-of-the-art are shown in \Tab{tbl:results}. \SEtwoBest{} was developed by \newcite{yarowsky2001johns} and achieved the best score on the English lexical sample task of \SEtwo{}{} with a F1 score of $64.2$. Their system utilized a rich feature space based on raw words, lemmas, POS tags, bag-of-words, bi-gram, and tri-gram collocations, etc. as inputs to an ensemble classifier. \SEthreeBest{} by \newcite{grozea2004finding} was the winner of the \SEthree{} lexical sample task with a F1 score of $72.9$. This system was based mainly on raw words, lemmas, and POS tags. 
These were used as inputs to a regularized least square classifier. \IMS{} is a more recent system, by \newcite{taghipour2015semi}, that uses separately trained word embeddings as input. However, it also relies on a rich set of other features including POS tags, collocations and surrounding words to achieve their reported result.

Our proposed model achieves the top score on \SEtwo{} and are tied with \IMS{} on \SEthree. Moreover, we see that \dropword{} consistently improves the results on both \SEtwo{} and \SEthree{}. Randomizing the order of the input words yields a substantially worse result, which provides evidence for our hypothesis that the order of the words are significant. We also see that the system effectively makes use of the information in the pre-trained word embeddings and that they are essential to the performance of our system on these datasets. \todo{Why is word embeddings so essential?}

\begin{table}[tbh]
\centering
\begin{tabular*}{0.9\textwidth}{@{\extracolsep{\fill} } lcr }
\toprule 
& \multicolumn{2}{c}{\bf F1 score} \\
\bf Method & \bf \SEtwo & \bf \SEthree \\
\midrule
\oursystem{} (our proposed model) & \bf 66.9 & \bf 73.4 \\
\midrule
\SEtwoBest & 64.2 & - \\
\SEthreeBest & - & 72.9 \\
\IMS & 66.2 & \bf 73.4 \\
\midrule
\oursystem{} without \dropword & 66.5 & 72.9 \\
\oursystem{} without \glove & 54.6 & 59.0  \\
\oursystem{}, \randomwordorder & 58.8 & 64.7 \\
\bottomrule
\end{tabular*}
\caption{Results for Senseval 2 and 3 on the English lexical sample task.}
\label{tbl:results}
\end{table}
\section{Conclusions \& future work}
We presented a BLSTM based model for WSD that was able to effectively exploit word order and achieve results on \emph{state-of-the-art} level, using no external resources or handcrafted features. 
As a consequence, the model is largely language independent and applicable to resource poor languages.
Further, the system was designed to generalize to full vocabulary WSD by sharing most of the parameters between words. 

For future work we would like to provide more empirical evidence for language independence by evaluating on several different languages, and do experiments on large vocabulary \emph{all words WSD}, where every word in a sentence is disambiguated. Further, we plan to experiment with unsupervised pre-training of the BLSTM, encouraged by the substantial improvement achieved by incorporating word embeddings. \todo{experiment to compare different window sizes}
%

\section*{Acknowledgments}
The authors would like to acknowledge the project \emph{Towards a knowledge-based culturomics} supported by a framework grant from the Swedish Research Council (2012--2016; dnr 2012-5738). Further, the work was partly funded by the \emph{BigData@Chalmers} initiative. Finally. the authors would like to thank the reviewers for their insightful and constructive feedback.

\bibliographystyle{acl}
\bibliography{coling2016}
\listoftodos
\end{document}